\documentclass[letterpaper]{article} % DO NOT CHANGE THIS
\usepackage{aaai25}  % DO NOT CHANGE THIS
\usepackage{times}  % DO NOT CHANGE THIS
\usepackage{helvet}  % DO NOT CHANGE THIS
\usepackage{courier}  % DO NOT CHANGE THIS
\usepackage[hyphens]{url}  % DO NOT CHANGE THIS
\usepackage{graphicx} % DO NOT CHANGE THIS
\usepackage{natbib}  % DO NOT CHANGE THIS AND DO NOT ADD ANY OPTIONS TO IT
\usepackage{caption} % DO NOT CHANGE THIS AND DO NOT ADD ANY OPTIONS TO IT
\usepackage{algorithm}
\usepackage{algorithmic}
\usepackage[algo2e]{algorithm2e}
\usepackage{threeparttable, booktabs}
\usepackage{multirow}
\usepackage{amsmath,amsfonts}
\newtheorem{myclaim}{Claim}
\newtheorem{mydefinition}{Definition}
\newtheorem{myproof}{Proof}

\usepackage{newfloat}
\usepackage{listings}
\DeclareCaptionStyle{ruled}{labelfont=normalfont,labelsep=colon,strut=off} % DO NOT CHANGE THIS
\floatstyle{ruled}
\newfloat{listing}{tb}{lst}{}
\floatname{listing}{Listing}
\title{First Activations Matter: Training-Free Methods for Dynamic Activation\\ in Large Language Models}
\author {
    Chi Ma\textsuperscript{\rm1},
    Mincong Huang\textsuperscript{\rm1},
    Ying Zhang\textsuperscript{\rm1},
    Chao Wang\textsuperscript{\rm1},
    Yujie Wang\textsuperscript{\rm1}\thanks{Corresponding author},\\
    Lei Yu\textsuperscript{\rm1},
    Chuan Liu\textsuperscript{\rm2},
    Wei LIn\textsuperscript{\rm2}
}
\affiliations {
    \textsuperscript{\rm 1}
    Meituan
    \textsuperscript{\rm 2}
    Individual \\ 
    \{wangyujie37@meituan.com\}
}

\begin{document}

\maketitle

\begin{abstract}
Dynamic activation (DA) techniques, such as DejaVu and MoEfication, have demonstrated their potential to significantly enhance the inference efficiency of large language models (LLMs). However, these techniques often rely on ReLU activation functions or require additional parameters and training to maintain performance. This paper introduces a training-free Threshold-based Dynamic Activation(TDA) method that leverage sequence information to exploit the inherent sparsity of models across various architectures. This method is designed to accelerate generation speed by 18-25\% without significantly compromising task performance, thereby addressing the limitations of existing DA techniques. Moreover, we delve into the root causes of LLM sparsity and theoretically analyze two of its critical features: history-related activation uncertainty and semantic-irrelevant activation inertia. Our comprehensive analyses not only provide a robust theoretical foundation for DA methods but also offer valuable insights to guide future research in optimizing LLMs for greater efficiency and effectiveness.
\end{abstract}

\section{Introduction}
Large Language Models (LLMs), such as LLaMA\cite{touvron2023llamaopenefficientfoundation, touvron2023llama2openfoundation}, Mistral\cite{jiang2023mistral7b}, Gemma\cite{gemmateam2024gemmaopenmodelsbased}, and the OPT\cite{zhang2022optopenpretrainedtransformer} series, have shown remarkable performance and in-context learning capabilities due to their extensive parameter counts. However, their substantial computational demands and latency during inference pose significant challenges. To address these issues, various techniques exploiting the inherent sparsity of LLMs have been proposed, aiming to reduce latency by minimizing the excessive activation of heads, neurons, and weights during inference.

Sparse activation techniques for LLMs can be categorized into \textit{static} and \textit{dynamic activation} methods. Static activation (SA) methods, such as pruning\cite{sun2024simpleeffectivepruningapproach, frantar2023sparsegptmassivelanguagemodels} and low-dimension projection\cite{ashkboos2024slicegptcompresslargelanguage}, reduce surplus weights in LLMs based on metrics like magnitude, applied once or progressively. These pruned structures remain fixed for all subsequent inputs and are fully activated during inference. However, SA has limitations: inactive weights cannot be restored after pruning, potentially degrading performance and in-context learning ability. Additionally, the iterative nature of SA requires substantial extra training, which may not yield proportional speedup enhancements.

On the other hand, dynamic activation (DA) offers adaptability by selectively activating certain heads or neurons during inference, thereby enhancing computational efficiency. This approach leverages the inherent sparsity in LLMs to optimize resource utilization. Existing researches on DA can be categorized in Table \ref{table: two types of da}.

\begin{table*}[h]
\begin{threeparttable}
    \centering
    \scalebox{0.75}{
    \begin{tabular}{c|p{7cm}p{3cm}p{2cm}p{4cm}}
    \toprule
    \textbf{DA Types} & \textbf{Definetions} & \textbf{Examples} & \textbf{Advantages} & \textbf{Current Limitations} \\
    \midrule
    \multirow{2}{*}{Training-Dependent DA} & Some leverage a \textit{predictor}, which is pre-trained using the model's training data, to dynamically identify essential activation neurons or experts during the model's forward. (Figure \ref{figure:Training-Dependent DA}) & DejaVu \cite{liu2023dejavucontextualsparsity}, MoEfication\cite{zhang2022moeficationtransformerfeedforwardlayers} & High Sparsity & Tend to underperform on models with non-ReLU activations(See Table~\ref{table: dejavu}) \\
    \cmidrule(lr){2-5}
    ~ & Others aim to reduce computational costs by employing multi-stage MoE-style training and introducing efficiency and separability loss penalties. & LTE \cite{zheng2024learnefficientbuildstructured} and D2DMoE \cite{szatkowski2024exploitingactivationsparsitydense} & High performance & Extra training required \\
    \midrule
    Training-Free DA & Employs pre-searched or pre-defined thresholds or sparsity levels to decide which neurons to retain or discard. Neurons with activation values falling below this bar are eliminated during current forward, thereby reducing computational overhead and latency.(Figure \ref{figure:Training-Free TDA}) & Griffin\cite{dong2024promptpromptedadaptivestructuredpruning} & Training-free for all model archs & Low performance \\
    \bottomrule
    \end{tabular}
    }
    \caption{Two types of DA methods}
    \label{table: two types of da}
\end{threeparttable}
\end{table*}

% \textbf{Training-Dependent Dynamic Activation}: 
% Some training-dependent DA methods leverage a \textit{router} to dynamically identify essential activation neurons or experts during the model's forward. This router is pre-trained using the model's historical data. (e.g.DejaVu\cite{liu2023dejavucontextualsparsity} and MoEfication\cite{zhang2022moeficationtransformerfeedforwardlayers}) Other training-dependent DA methods aim to reduce computational costs by employing multi-stage training and introducing efficiency and separability loss penalties. (e.g.LTE \cite{zheng2024learnefficientbuildstructured} and D2DMoE \cite{szatkowski2024exploitingactivationsparsitydense})
% \textbf{Training-Free Threshold Dynamic Activation (TDA)}:
% TDA employs pre-searched or pre-defined thresholds or sparsity levels to decide which neurons to retain or discard. Neurons with activation values falling below this bar are eliminated during current forward, thereby reducing computational overhead and latency. 

\begin{table}[]
\centering
\scalebox{0.8}{
\begin{tabular}{r|cccc}
\toprule
\multicolumn{1}{c}{\textbf{}}          & \textbf{MMLU} & \textbf{TruthfulQA} & \textbf{Winogrande} & \textbf{GSM8K} \\
\midrule
\multicolumn{1}{c}{\textbf{LLaMA2-7B}} & 45.83         & 61.04               & 74.11               & 13.95          \\
\midrule
TT      & 45.62 & 60.66 & 73.88 & 13.65 \\
Griffin & 43.59 & 59.26 & 73.21 & 12.31 \\
TDA     & 44.83 & 60.45 & 73.53 & 13.18 \\
\midrule
DejaVu  & 27.02 & 51.12 & 50.2  & 7.22  \\
\bottomrule
\end{tabular}
}
\caption{Dejavu tends to underperform on models with non-ReLU activations.}
\label{table: dejavu}
\end{table}

\begin{figure*}[htbp]
    \begin{minipage}[t]{0.5\linewidth}
        \centering
        \includegraphics[height=2.7in]{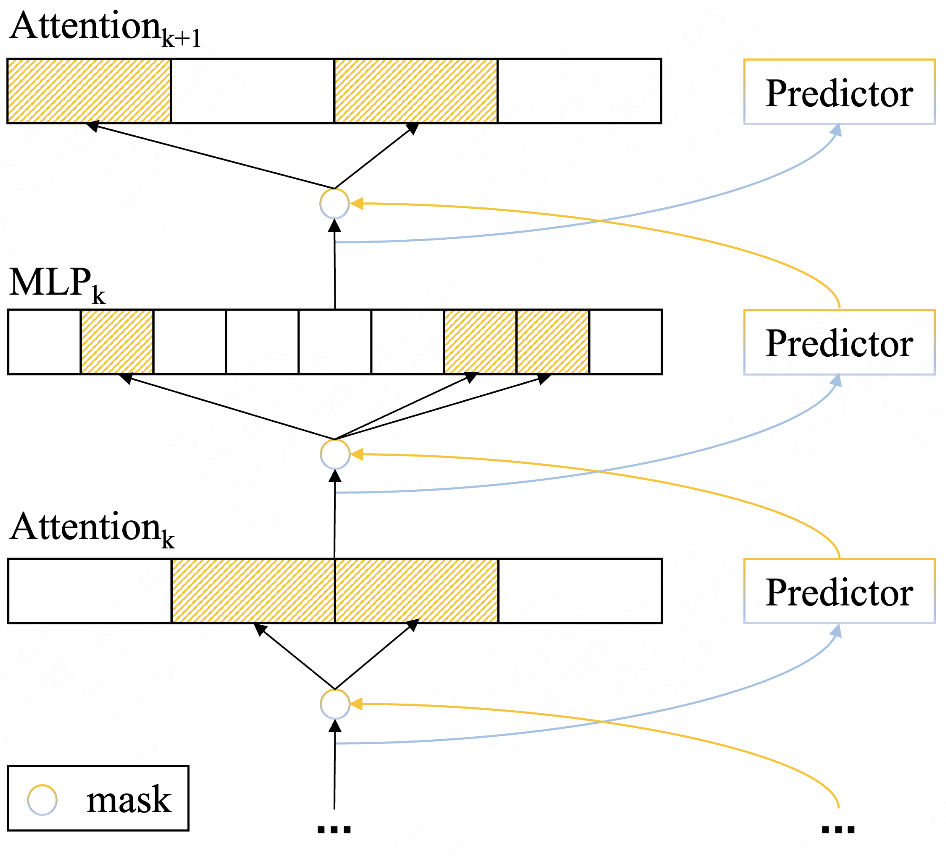}
        \caption{Training-Dependent DA}
        \label{figure:Training-Dependent DA}
    \end{minipage}
    \begin{minipage}[t]{0.5\linewidth}
        \centering
        \includegraphics[height=2.7in]{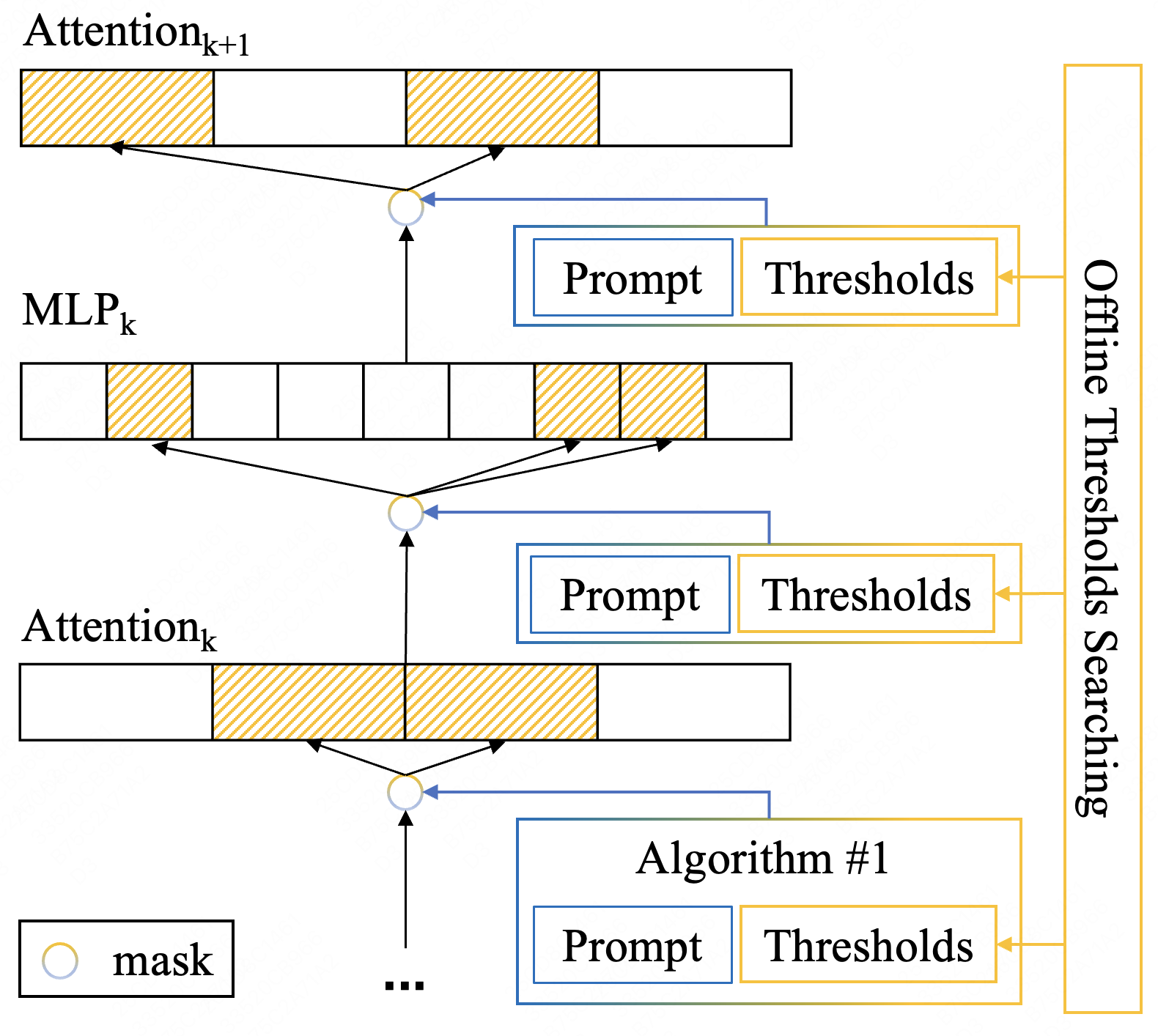}
        \caption{Training-Free TDA}
        \label{figure:Training-Free TDA}
    \end{minipage}%
\end{figure*}

Our approach builds upon the creative idea presented by ReLU$^2$\cite{zhang2024relu2winsdiscoveringefficient} and Griffin\cite{dong2024promptpromptedadaptivestructuredpruning}, where they proposed threshold calculation, truncation and sequential flocking. From Figure \ref{figure:Training-Free TDA} we can see that, unlike training-dependent DA methods in Figure \ref{figure:Training-Dependent DA} which use a pre-trained predictor directly for inference, the proposed method accelerates generation by calculating the L2 Norm of the up and gate projections in the prompt section to obtain a mask. This approach can improve generation time by 18-25\% with minimal loss in model accuracy.

Despite significant progress, current research on DA still lacks a comprehensive theoretical framework that examines its phenomena and underlying mechanisms. In this paper, we also present a mathematical explanation for the causes of DA and analyze two of its key characteristics: history-related activation uncertainty and semantic-irrelevant activation inertia.

The key contributions of this paper are:
\begin{enumerate}
    \item Propose TDA, which significantly reduces generation latency with minimal impact on model performance.
    \item Provide a mathematical explanation for DA and its relationship with the ReLU activation function.
    \item Identify history-related activation uncertainty(in Section \ref{sec: his}) in dynamic activation, explaining why previous DA methods (e.g. DejaVu) fail in models with non-ReLU activation functions.
    \item Conduct a detailed analysis of semantic-irrelevant activation inertia(in Section \ref{sec: sem}) in DA, elucidating the mechanism of TDA that leverages sequential information in models across various architectures and activation functions.
\end{enumerate}

The remainder of the paper is organized as follows: Section \ref{section:related works} reviews related works. In Section \ref{section:Preliminaries}, we introduce our theoretical analysis. Section \ref{section:Methodology} presents the TDA methods, followed by extensive experiments in Section \ref{section:experiments}. Finally, conclusions are drawn in Section \ref{section:conclusion}.

\section{Related Works}\label{section:related works}
\subsection{Inherent Sparsity in LLMs}
In Large Language Models (LLMs), \textit{inherent sparsity} refers to the excessive activation of neurons during tasks, leading to inefficiency and wasted resources\cite{bommasani2022opportunitiesrisksfoundationmodels, yuan2024llminferenceunveiledsurvey}. Studies\cite{liu2023modelbasedcontrolsparseneural} show that dense neural networks often display surplus activation. Treating sparsity as a continuous process can optimize model architecture holistically. The Lottery Hypothesis\cite{frankle2019lotterytickethypothesisfinding, malach2020provinglotterytickethypothesis} highlights pruning techniques to remove unnecessary connections and leverage inherent sparsity.

Other research\cite{shazeer2017outrageouslylargeneuralnetworks} addresses this with \textit{sparse activation} using a sparsely-gated mixture-of-experts (MoE) layer, increasing model capacity while reducing computational costs. MC-SMoE\cite{li2024mergecompressdemystifyefficient} further optimizes MoEs by merging and low-rank decomposition of redundant experts, guided by the router's information.

\subsection{Dynamic Activation}
\subsubsection{Training-Dependent DA with ReLU}
Research\cite{liu2023modelbasedcontrolsparseneural, mirzadeh2023relustrikesbackexploiting} highlights the ability of the ReLU activation function to introduce activation sparsity and proposes the concept of dynamic activation.
DejaVu\cite{liu2023dejavucontextualsparsity} demonstrates that the sparsity induced by ReLU can be predicted, leading to the first viable training-dependent neuron-level DA scheme by adding a pre-trained two-layer linear router before FFN block. On the OPT series, DejaVu achieves a 2-6x acceleration in inference latency at 75\% sparsity.

Building on the DejaVu idea, ReLU$^2$\cite{zhang2024relu2winsdiscoveringefficient} proposed a new ReLU$^2$ activation function that could attain nearly 70\% sparsity with minimal performance loss.
ProSparse\cite{song2024prosparseintroducingenhancingintrinsic} introduces a practical DA inference framework and achieves only a 1\% increase in perplexity at approximately 80\% sparsity by replacing the activation function and continuing to induce sparsity.

Related works of training-dependent DA in MoEs can be seen in Appendix~\ref{app:training-dependent da in moes}.

\subsubsection{Training-free DA}
As the first training-free method, Griffin\cite{dong2024promptpromptedadaptivestructuredpruning} selects neurons by leveraging the sparse activation pattern known as \textit{flocking} at the sequence level in LLMs. This approach halves the computational requirements for the generation phase, thereby reducing latency with a 1-3\% performance decrease.

\section{Preliminaries}\label{section:Preliminaries}
Section~\ref{section:related works} reviewed the literature pertinent to the inherent sparsity of LLMs and dynamic activation. This section begins by presenting a theoretical analysis of the causes of sparsity and the limitations of previous DA methods. Additionally, it underscores the necessity of incorporating sequence information into the DA method by examining two key features of DA.

\subsection{Inherent Sparsity of LLMs}\label{sec: unveiling}
Following the literature\cite{li2023lazyneuronphenomenonemergence}, we can demonstrate through the subsequent derivation how sparsity arises and why SwiGLU cannot produce greater sparsity than ReLU.

\begin{myclaim}
\label{claim:cause of sparsity}
    Any training algorithm based on negative gradient directions tends to \textit{reduce the magnitude} of \textit{positive activation}, since it will lead to a smaller training loss, and thus causes \textit{sparsity}.
\end{myclaim}

\begin{mydefinition}
Assuming a neural network as in Equation~\ref{eq: initiate}:
\begin{equation}
    f(x)=\boldsymbol{V}\sigma (p(\boldsymbol{x};\boldsymbol{\theta}))
\label{eq: initiate}
\end{equation}
,where \(\boldsymbol{V} = [v_1, ..., v_{d_{ff}}]\) is network parameter for the last layer drawn from a random distribution, \(\sigma()\) is the SwiGLU activation function, and \(p(\boldsymbol{x};\boldsymbol{\theta})\) denotes all other layers with parameter \(\theta\). We write \(p = p(\boldsymbol{x};\boldsymbol{\theta})\) for simplicity. 
\end{mydefinition}

\begin{mydefinition}
Consider the cross-entropy (CE) loss with function \(\ell_{CE}(f(\boldsymbol{x}),\boldsymbol{y})\), where \(\boldsymbol{y}\) is an arbitrary vector that sums up to one and independent of \(\boldsymbol{V}\).

Assume that the entries of \(\boldsymbol{V}\) are drawn from independent distributions, the probability of any entry of \(\boldsymbol{V}\) being 0 is less than 1, and \(E[\boldsymbol{V}] = 0\) . 

If there exist an \(i^*\) such that \(p_{i^*} > 0\), then we have Equation \ref{eq: lianshiqiudao}:
\begin{equation}
    \frac{\partial \ell}{\partial p_{i*}} =\left \langle \frac{\partial \ell}{\partial f}, \frac{\partial f}{\partial p_{i*}} \right \rangle=\left \langle \frac{\partial \ell}{\partial f},v_{i^*} \right \rangle  
\label{eq: lianshiqiudao}
\end{equation}
\end{mydefinition}

\begin{myproof}
Substituting CE loss function into Equation \ref{eq: lianshiqiudao} yields Equation \ref{eq: lianshiqiudao first term}:
\begin{equation}
\begin{aligned}
    \frac{\partial \ell_{CE}}{\partial f}
    &=\frac{exp(f(x))}{\left \langle exp(f(x)),\boldsymbol{1} \right \rangle }-y\\
    &=\frac{exp( {\textstyle \sum_{i}\sigma(p_i)\cdot \boldsymbol{v_i}})}{\left \langle exp( {\textstyle \sum_{i}\sigma(p_i)\cdot \boldsymbol{v_i}}), \boldsymbol{1}\right \rangle}-y
\end{aligned}
\label{eq: lianshiqiudao first term}
\end{equation}

By substituting Equation \ref{eq: lianshiqiudao first term} back into Equation \ref{eq: lianshiqiudao}, we can obtain Equation \ref{eq: lianshiqiudao second term}:
\begin{equation}\footnotesize
\begin{aligned}
    \frac{\partial \ell_{CE}}{\partial p_{i^*}}
    &=\frac{\left \langle exp(\sum_{i}\sigma(p_i) \cdot \boldsymbol{v_i} ),\boldsymbol{v_{i^*}} \right \rangle }{\left \langle exp(\sum_{i}\sigma(p_i) \cdot \boldsymbol{v_i} ),\boldsymbol{1} \right \rangle }- \left \langle \boldsymbol{v_{i^*}},y \right \rangle 
\end{aligned}
\label{eq: lianshiqiudao second term}
\end{equation} 

Similar to literature\cite{li2023lazyneuronphenomenonemergence}, we also have \(\mathrm {E}[\frac{\partial \ell_{CE}}{\partial p_{i^*}}] > 0\) holds true since the expectation of \textbf{V} is zero and the transformation of the activation function does not change the non-negative property of the loss expectations. Detailed derivation process of Equation~\ref{eq: loss expectation} can be found in Appendix~\ref{app: proof of claim}.
\begin{equation}\footnotesize
\begin{aligned}
    \mathrm {E}[\frac{C_1V \cdot exp(pV)}{C_2~exp(pV)+C_3}] =\mathrm {E}[\frac{C_1V}{C_2 + C_3exp(-pV)} ]
\end{aligned}
\label{eq: loss expectation}
\end{equation} 
\end{myproof}

The first term on the right-hand side(RHS) of the loss function(in Equation \ref{eq: lianshiqiudao second term})'s expectation can be simplified to the form of Equation \ref{eq: loss expectation}, while the expectation of the second term on the RHS is zero. With respect to \(p_{i^*}^0 < p_{i^*}^1\), we have Equation \ref{eq: loss expectation} demonstrates that when the activation function is switched from ReLU to SwiGLU, the expected value of the loss function will decrease. 

That is to say: if there exist an \(i^*\) such that \(p_{i^*} > 0\), the gradient of CE loss with respect to any positive activation \(p_{i^*} > 0\) is positive in expectation. 

Therefore, Claim 1 is proved. And ReLU activation function will cause a bigger magnitude reduction that SwiGLU in this process.

\subsection{History-related Activation Uncertainty}\label{sec: his}
In Section \ref{sec: unveiling}, this paper theoretically deduces the root causes of inherent sparsity and explores how non-ReLU activation functions might mitigate it. The literature \cite{georgiadis2019acceleratingconvolutionalneuralnetworks, kurtz2020pmlr, zhu2023cvprw} has also highlighted that the current level of sparsity is insufficient to fully unlock the performance of DA methods, especially for non-ReLU activated models\cite{ma2024dynamicactivationpitfallsllama, dong2024promptpromptedadaptivestructuredpruning}. In Sections \ref{sec: his} and \ref{sec: sem}, we analyze two key features of dynamic activation and, from this, illustrate the rationale behind the proposed TDA method.

\begin{myclaim}
\label{claim:failure of training-dependent DA}
    The failure of training-dependent DA on non-ReLU activated models is linked to the shifts in weight importance under different history inputs.
\end{myclaim}

This is to say: a predictor trained on different historical activation data may find it difficult to accurately identify the weights that are most crucial for the current input.

Similarly, we assume the presence of a ReLU-activated model as described in Equation \ref{eq: initiate}. And:
\begin{mydefinition}\label{definition1}
the simplified loss of current input token \(x_i\) can be described as (Equation \ref{eq: model loss}):
\begin{equation}\footnotesize
\begin{aligned}
    L_i=(\frac{\partial f}{\partial x_i}\mathrm{d}x_i + \frac{\partial f}{\partial \mathbf{\theta}_i } \mathrm{d}\mathbf{\theta}_i)^T(\frac{\partial f}{\partial x_i}\mathrm{d}x_i + \frac{\partial f}{\partial \mathbf{\theta}_i } \mathrm{d}\mathbf{\theta}_i)
\end{aligned}
\label{eq: model loss}
\end{equation}
\end{mydefinition}

\begin{myproof}
Based on Equation~\ref{eq: model loss} in Definition~\ref{definition1}, weight change sensitivity (gradients) in model training is as Equation \ref{eq: output qiudao}:
\begin{equation}
\begin{aligned}
    \frac{\partial L_i}{\partial \mathrm{d}\mathbf{\theta}_i} = 2(\frac{\partial f}{\partial x_i}\mathrm{d}x_i + \frac{\partial f}{\partial \mathbf{\theta}_i } \mathrm{d}\mathbf{\theta}_i)\frac{\partial f}{\partial \mathbf{\theta}_i}
\end{aligned}
\label{eq: output qiudao}
\end{equation}

By summing gradients, we have Equation \ref{eq: gradient sum}:
\begin{equation}
\begin{aligned}
    \nabla _{\mathrm{d}\theta_i}L 
    &= \sum_{i}2(\frac{\partial f}{\partial x_i}\mathrm{d}x_i + \frac{\partial f}{\partial \mathbf{\theta}_i } \mathrm{d}\mathbf{\theta}_i)\frac{\partial f}{\partial \mathbf{\theta}_i}\\
    &= \nabla _{\mathrm{d}\theta_i}L_i+\sum_{j=0:i-1}\nabla _{\mathrm{d}\theta_j}L_{j}
\end{aligned}
\label{eq: gradient sum}
\end{equation}

And the importance of model weights can be described in Equation \ref{eq: weight importance}:
\begin{equation}
\begin{aligned}
    \Theta_i 
    &= \sum_{i}|V \cdot \nabla _{\mathrm{d}\theta_i}L_i| \\
    &=|V|\cdot \sum_{i}|\nabla _{\mathrm{d}\theta_i}L_i|\\
    &=|V |\cdot (\nabla _{\mathrm{d}\theta_i}L_i+\sum_{j=0:i-1}\nabla _{\mathrm{d}\theta_j}L_{j})\\
    &= |V |\cdot \nabla _{\mathrm{d}\theta_i}L_i+\Theta_{i-1}
\end{aligned}
\label{eq: weight importance}
\end{equation}
, which means weight importance of a model are not only related to current input along the direction of \(\theta\), but also to the cumulative gradient information from all previous data. 
\end{myproof}

For models utilizing ReLU activation, Equation \ref{eq: weight importance} can be simplified to the sum of the weights corresponding to positive inputs, which linearly correlates with the magnitude of the current weights themselves. However, for models employing non-ReLU activations, the importance of current weights contains information from all previous tokens.

\subsection{Semantic-irrelevant Activation Inertia}\label{sec: sem}
By using simplified loss function, Section \ref{sec: his} demonstrated that models with non-ReLU activation need historical information to accurately decide which neurons will be activated. This section reveals that:
\begin{myclaim}
    Activation inertia is irrelevant with semantic information of the current token.
\end{myclaim}

This is intuitive since historical information is significantly influenced by the Heavy Hitter(\(H_2\)), and the occurrence of \(H_2\) is not related to semantics\cite{sun2024massiveactivationslargelanguage}.

\begin{mydefinition}
Following literature\cite{zhang2023h2oheavyhitteroracleefficient} we have:
\begin{itemize}
    \item \(H_2: S^*\subset [m]\), where m denotes the length, and
    \item \(k = |S^*|,~\tau \in (0, 1)\) denotes a threshold, and
    \item \(\alpha \in (0,1)\) denote a fraction of mass (larger than \(\tau\)) outside \(S^*\).
\end{itemize} 
\end{mydefinition}

\begin{myproof}
    It is natural that attention with \(H_2\) is a \((\alpha, \tau, k)\)-good mapping since for all \(x \in \mathbb{R}^d\), \(S^* \subset supp_\tau(Att(x))\), and \(|supp_\tau(Att(x))\setminus S^*| \le \alpha \cdot k\). Then we have \(S^* \subseteq \cap_{i\in[n]}supp_\tau(x_i)\), and \(|(\cup_{i\in[n]}supp_\tau(Att(x)))\setminus S^*| \le \alpha k n\) for \(x_i\) draw from \((\alpha, \tau, k)\)-good distribution uniformly at random. That is to say, \(H_2\) in a sequence significantly decides the activation pattern.
\end{myproof}

In summary, Section~\ref{sec: his} and \ref{sec: sem} theoretically demonstrate that for neurons activated by the current token, the influence of preceding tokens in the same sequence far outweighs the semantic influence of the current token.

In short:
\begin{myclaim}\label{final claim}
    Within a sequence, neuron activation pattern is more influenced by activation inertia than by the semantics of the current token.
\end{myclaim}

\subsection{A Closer Look at Activation Inertia}
\begin{figure*}[h]
    \begin{minipage}[t]{0.5\linewidth}
        \centering
        \includegraphics[height=2in]{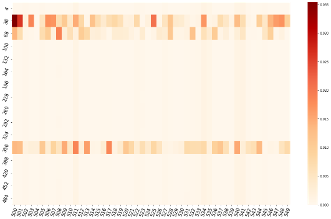}
        \caption{Active pattern of 16 tokens separately}
        \label{figure:oneword}
    \end{minipage}
    \begin{minipage}[t]{0.5\linewidth}
        \centering
        \includegraphics[height=2in]{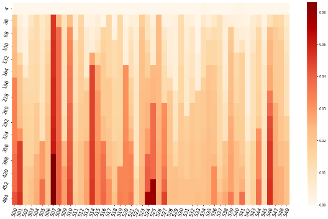}
        \caption{Active pattern of these 16 tokens as a sentence}
        \label{figure:onewordseq}
    \end{minipage}
        \begin{minipage}[t]{0.5\linewidth}
        \centering
        \includegraphics[height=2in]{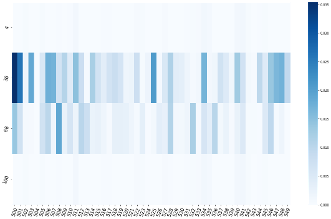}
        \caption{Active pattern of 4 random tokens separately}
        \label{figure:onewords}
    \end{minipage}
        \begin{minipage}[t]{0.5\linewidth}
        \centering
        \includegraphics[height=2in]{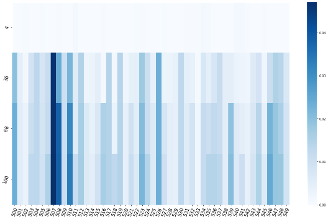}
        \caption{Active pattern of these 4 random tokens\protect\\ as a sequence}
        \label{figure:onewordseqs}
    \end{minipage}
\end{figure*}

To further investigate Claim~\ref{final claim} in a more intuitive and detailed manner, we extracted the first 16 tokens from the first entry of the XSum 1-shot dataset to generate Figures \ref{figure:oneword} to \ref{figure:onewordseqs}. In these figures, the horizontal axis represents the neuron indices, while the vertical axis represents the word indices.

Figures \ref{figure:oneword} to \ref{figure:onewordseqs} show that in a sequence, neurons activated by each token are influenced by preceding tokens, especially for randomly selected tokens, indicating inertia in neuron activation.

For a comprehensive analysis on larger datasets, see Appendix~\ref{app: larger scale}.

The mechanism behind this phenomenon needs further investigation. When processed as a sequence, activated neurons consider all tokens, suggesting that inertia may be due to preceding tokens rather than the current token.

To eliminate sequence information influence, we selected specific samples and conducted a similarity analysis of their activation patterns. Details in Table~\ref{table: detailed 13 samples} in Appendix~\ref{app: detailed samples}.

The aforementioned facts shows that models can discriminates semantic difference and thus collectively validate Claim~\ref{final claim}.

\section{Methodology}\label{section:Methodology}
Using our insight on the importance of sequence information in activation, we introduce our TDA methods as a simple and training-free method for dynamic activation. Shortly, we selectively activate neurons in generation phase based on previous sequential information.

Threshold truncation(TT) proposed by ReLU$^2$\cite{zhang2024relu2winsdiscoveringefficient} already leverages an offline-searched thresholds to determine which LLMs heads or neurons under different inputs should be retained. TT offers the advantage of having minimal impact on the model's performance.

However, a notable drawback is its dependency on the online computation of decision matrices of neurons. For example, in LLaMA-2 models, TT requires calculating both the up and gate projections for each token, followed by computing their L2 norm as the decision metric. This means that only the calculation for the down projection is reduced by less than 50\%, while the extra L2 norm calculation is still required. Consequently, the speedup achieved is not significant.

Following TT, our TDA method is detailed in Algorithm 1. Compared with TT, TDA follows its offline threshold search but significantly reduces online computation by reusing the activation patterns of the prompt section, as explained in the theoretical framework of the Section.

During the prefill phase of TDA, it iterates through each layer of the model, computes feedforward activations $A$ and normalizes these activations to obtain relative magnitudes $R$ in accordance with TT. Then, TDA selects neurons above the layer-wise threshold $thres[i]$ as the mask $mask$, and stores this mask in \texttt{mask\_array}. In the generation phase, for each step in the generation length, it iterates through each layer, retrieves the corresponding mask $mask$ from \texttt{mask\_array}, uses this mask to compute the sliced feedforward network $\tilde{FF}$ to compute hidden states $H$.

From Algorithm 1, we can see that TDA method employs different thresholds for each layer of the model, allowing the number of activated neurons to vary across layers. Compared to the static top-k approach used by Griffin, this method provides a significant advantage in maintaining model performance.

\begin{algorithm}[h]
\caption{Threshold Dynamic Activation}
\LinesNumbered
\KwIn{Model parameters $\theta$, dataset $p$, layer-wise threshold $thres$, generation length $len$}
\KwOut{Generated text $g$}

Initialize model with parameters $\theta$;

\For{each sample in dataset $p$}{
    \tcp{Initialize an array to store masks for each layer}
    \texttt{mask\_array} $\gets$ [ ]

    \tcp{Prefill phase}
    \For{$i \gets 1$ \textbf{to} num\_layers}{
        Compute feedforward activations $A$;
        
        Normalize activations to get relative magnitudes $R$;
        
        Select neurons above \texttt{thres[i]} as mask $mask$;
        
        \texttt{mask\_array[i]} $\gets$ $mask$;
        }

    \tcp{Generation phase}
    \For{each step in $len$}{
        \For{$j \gets 1$ \textbf{to} num\_layers}{
            $mask$ $\gets$ \texttt{mask\_array[j]};
            
            Use $mask$ to compute sliced FFN $\tilde{FF}$;
            
            Compute hidden states $H$ using $\tilde{FF}$;
            
            Predict next token probabilities using $H$;
            
            Sample next token $t$ from probabilities;
            
            Append token $t$ to generated sequence $g$;
            }
        }
    Return generated sequence $g$;
    }

\label{algo}
\end{algorithm}

% \begin{algorithm}[h]
% \caption{Threshold Dynamic Activation}%算法名字
% \LinesNumbered %要求显示行号
% \KwIn{Model parameters $\theta$, dataset $p$, layer-wise threshold $thres$, generation length $len$}%输入参数
% \KwOut{generated text $g$}%输出  

% Initialize model with parameters $\theta$;

% \ForEach{line in dataset $p$}{

% Normalize prompt $p$ to token embeddings;

% \tcp{Prefill phase}\label{cmt}
% \ForEach{layer in model}{
%     Compute feedforward activations $A$;
    
%     Normalize activations to get relative magnitudes $R$;

%     Select neurons above $thres$ as Mask $mask$;

%     \tcp{Generation phase}\label{cmt}
%     \For{each step in $len$}{
%         Use $mask$ to compute sliced FFN $\tilde{FF}$;
        
%         Compute sequence of hidden states $H$ using $\tilde{FF}$;
        
%         Predict next token probabilities using $H$;
        
%         Sample next token $t$ from probabilities;
        
%         Append token $t$ to generated sequence $g$;
% }    
% }
% }

% Return generated sequence $g$;
% \label{algo}
% \end{algorithm}

For the calculation principles of the layer-wise threshold, please refer to the Appendix~\ref{app: cett}.

\section{Experiments}\label{section:experiments}
\subsection{Setups}
Our approach, along with the baseline models, is implemented using the PyTorch framework, and we leverage the Hugging Face Transformers library for model and dataset management. Our experiments are powered by 1 NVIDIA A100 GPUs with 80 GB of memory. Adhering to the methodologies outlined in Section \ref{section:Methodology}, we sequentially applied our methods for each Transformer layers, which reduces inference latency while preserving model performance. All experiments are conducted in a single phase, without any post-training or fine-tuning stages.

\subsubsection{Models, Datasets.} 
In this paper, we conducted a comprehensive series of experiments using the OPT-350M, OPT-2.7B, Gemma-2B, LLaMA-2-7B and LLaMA-3-8B and Mistral-7B models. These models represent a significant advancement in language modeling capabilities, providing a spectrum of scales to meet various computational needs and performance benchmarks.

Following Griffin\cite{dong2024promptpromptedadaptivestructuredpruning}, we conduct evaluations on a variety of models across multiple generation and classification tasks. For generation tasks, we focus on XSum\cite{narayan2018dontdetailsjustsummary}, CNN/DailyMail\cite{nallapati2016abstractivetextsummarizationusing}, COQA\cite{reddy2019coqaconversationalquestionanswering}, and QASPER\cite{shaham2022scrollsstandardizedcomparisonlong}. For classification tasks, our evaluation includes HellaSwag\cite{zellers2019hellaswagmachinereallyfinish}, PIQA\cite{bisk2019piqareasoningphysicalcommonsense}, COPA\cite{roemmele2011choice}, ARC-Challenge\cite{clark2018thinksolvedquestionanswering}, and BoolQ\cite{clark2019boolqexploringsurprisingdifficulty}. Except for XSum and CNN/DailyMail, our experiments utilize the LM Evaluation Harness\cite{eval-harness}. 

\subsubsection{Baselines.}
Besides comparing against the original LLM, we also evaluate TDA in relation to Griffin and the TT methods introduced by ReLU$^2$. Unless specified otherwise, each technique is applied in a layer-wise manner, enhancing scalability even when dealing with exceptionally large models. TT has same performance with TDA, therefore we only evaluate its generation phase latency. For previous DA methods like DejaVu, we did not select it as a baseline for comparison in subsequent experiments. The reason is that DejaVu fails on models with non-ReLU activations (see Table~\ref{table: dejavu}), making it not comparable to the method proposed in this paper.

\subsubsection{Sparsity.}
In our evaluation, we especially focus on the MLP blocks of LLM models, which constitute approximately 67\% of the parameters of model's two main blocks, making them a crucial target for dynamic activation. We investigate two types of DA: Griffin and TDA with 50\% of sparsity, which facilitates a more fair comparison and deeper understanding of how different DA methods affect the performance of LLMs. 

\subsection{Performance}
Table \ref{table: performance} delineates the performance differences between the Griffin and TDA methods across various generation and classification tasks. Metrics such as Accuracy (Acc), Rouge-1, and F1 scores were measured across various datasets.

\begin{table*}[]
\centering
\begin{tabular}{c|ccccc|cc|cc}
\toprule
 &
  \multicolumn{5}{c}{\textbf{Acc}} &
  \multicolumn{2}{c}{\textbf{Rouge-1}} &
  \multicolumn{2}{c}{\textbf{F1}} \\
  \midrule
\textbf{Models} &
  \textbf{Hellaswag} &
  \textbf{Piqa} &
  \textbf{Copa} &
  \textbf{Arc-c} &
  \textbf{Boolq} &
  \textbf{Xsum} &
  \textbf{Cnn} &
  \textbf{Coqa} &
  \textbf{Qasper} \\
  \midrule
\textbf{OPT-350M}   & 32.06 & 64.64 & 72.00 & 21.33 & 41.01 & 12.89 & 14.82 & 33.39 & 3.34  \\
Griffin    & 30.52 & 62.46 & 69.00 & 20.24 & 39.71 & 10.59 & 13.32 & 31.89 & 2.14  \\
TDA        & 32.00 & 64.04 & 72.00 & 20.73 & 40.76 & 11.23 & 13.47 & 32.24 & 2.45  \\
\midrule
\textbf{OPT-2.7B}   & 45.86 & 73.78 & 77.00 & 60.77 & 66.79 & 18.43 & 22.24 & 64.41 & 7.85  \\
Griffin    & 43.76 & 71.84 & 76.00 & 58.21 & 65.92 & 17.43 & 20.74 & 62.91 & 6.85  \\
TDA        & 45.74 & 73.18 & 76.00 & 58.42 & 66.19 & 17.86 & 21.33 & 64.05 & 7.70  \\
\midrule
\textbf{Gemma-2B}   & 71.40 & 77.30 & 83.00 & 42.10 & 69.40 & 15.69 & 23.32 & 72.03 & 12.46 \\
Griffin    & 70.03 & 76.34 & 82.00 & 41.19 & 68.42 & 14.69 & 22.18 & 71.78 & 11.83 \\
TDA        & 70.85 & 76.21 & 82.00 & 41.12 & 68.21 & 15.32 & 22.51 & 72.45 & 12.33 \\
\midrule
\textbf{LLaMA-2-7B} & 57.16 & 78.07 & 87.00 & 43.34 & 77.71 & 27.15 & 10.08 & 77.35 & 26.31 \\
Griffin    & 56.66 & 76.57 & 85.00 & 41.84 & 76.21 & 26.65 & 8.58  & 75.85 & 25.81 \\
TDA        & 56.86 & 77.67 & 86.00 & 42.84 & 77.51 & 26.85 & 9.98  & 76.95 & 26.11 \\
\midrule
\textbf{LLaMA-3-8B} & 62.53 & 81.85 & 93.00 & 46.29 & 80.76 & 29.62 & 12.21 & 82.92 & 28.86 \\
Griffin    & 62.03 & 80.35 & 91.00 & 43.79 & 78.26 & 27.12 & 11.71 & 82.42 & 27.36 \\
TDA        & 62.31 & 81.40 & 92.00 & 45.79 & 80.39 & 29.47 & 11.93 & 82.57 & 28.37 \\
\midrule
\textbf{Mistral-7B} & 61.21 & 80.58 & 92.00 & 50.43 & 83.61 & 28.67 & 28.00 & 80.70 & 24.56 \\
Griffin    & 59.71 & 79.08 & 92.00 & 47.43 & 82.11 & 27.17 & 26.50 & 78.20 & 22.06 \\
TDA        & 59.32 & 79.21 & 92.00 & 49.24 & 83.14 & 28.35 & 27.53 & 80.55 & 24.07 \\
\bottomrule
\end{tabular}
\caption{Generation and classification performance across various model architectures.}
\label{table: performance}
\end{table*}

Overall, the performance of Griffin and TDA are subtle compared with dense models, but TDA continues outperforming Griffin in all tasks.

For instance, with the OPT-350M model, Griffin achieves slightly lower accuracy on Hellaswag (30.52) compared to TDA (32.00), and similar trends are observed in datasets like Piqa, Copa, and Boolq. Similarly, in generation tasks such as Xsum and CNN, Griffin tends to have lower Rouge-1 and F1 scores compared to TDA, indicating that TDA might be more effective in both classification and generation scenarios.

As the model size increases, the differences between Griffin and TDA become more pronounced. For example, with the LLaMA-3-8B model, TDA outperforms Griffin in most tasks, including Hellaswag, Piqa, and Copa, while also achieving higher Rouge-1 and F1 scores in generation tasks. For the Mistral-7B model, TDA generally has a slight edge over Griffin in both classification and generation tasks, suggesting that TDA might offer better overall performance as model complexity increases.

In summary, while both Griffin and TDA variants perform comparably, TDA often has a slight advantage in both classification and generation tasks, especially as model size increases. This is because Griffin consistently selects a fixed top-k, which may discard some important neurons. In contrast, the TDA method proposed in this paper employs a threshold-based dynamic top-k, providing greater adaptability.

\subsection{Efficiency}
Table \ref{table: latency} provides a comparative analysis of the generation latency for various models on a single NVIDIA A100 GPU, using a batch size of 1 and models implemented in FP16 precision via Hugging Face. The evaluated models include OPT-2.7B, Gemma-2B, LLaMA-2-7B, and Mistral-7B, with latency measured across different configurations: Dense, Griffin, TT, and TDA. Both the prompt length and the generated new token length are set to 1024, and the sparsity of Griffin and TDA is 50\%. The unit of reported numbers in Table \ref{table: latency} is seconds.

\begin{table*}[h]
\centering
\scalebox{1.0}{
\begin{tabular}{c|ccc|c}
\toprule
\textbf{Models} & \textbf{Dense} & \textbf{TT} & \textbf{Griffin} & \textbf{TDA}   \\
\midrule
OPT-2.7B    & 32.95 & 33.52 & 26.96(22.22\%$\downarrow$) & 27.77(18.65\%$\downarrow$)  \\
Gemma-2B    & 30.17 & 30.16 & 23.92(26.13\%$\downarrow$) & 24.06(25.39\%$\downarrow$)  \\
LLaMA-2-7B  & 80.31 & 78.88 & 64.32(24.86\%$\downarrow$) & 66.25(21.22\%$\downarrow$)  \\
Mistral-7B  & 79.28 & 76.26 & 63.26(25.32\%$\downarrow$) & 64.94(22.08\%$\downarrow$)  \\
\bottomrule
\end{tabular}
}
\caption{Generation phase latency(s).}
\label{table: latency}
\end{table*}

The results demonstrate that the TDA method consistently reduces generation latency compared to the dense configuration across all evaluated models. As shown in Table~\ref{table: latency}, both Griffin and TDA offer similar speedups, ranging from 18-25\%, whereas TT maintains a similar generation latency to dense models.

Overall, while Griffin is slightly faster in terms of latency, TDA offers greater advantages in maintaining model performance, and the TT method's latency is significantly higher than the other two methods. These results underscore the efficiency of TDA in accelerating generation speed without significantly compromising task performance, making it a practical and effective solution for optimizing large language models. Overall, TDA not only improves the efficiency of LLMs but also ensures that they remain precise, thereby broadening their potential use cases.

\section{Conclusion}\label{section:conclusion}
In summary, this paper introduces a training-free Threshold Dynamic Activation (TDA) method designed to enhance the inference efficiency of large language models (LLMs). By leveraging sequence information to exploit the inherent sparsity of models across various architectures, TDA achieves a 18-25\% improvement in generation speed without significantly affecting task performance. Unlike existing dynamic activation (DA) techniques such as DejaVu and MoEfication, which often require specific activation functions or additional structures and training, TDA offers a more practical and straightforward solution, addressing the limitations of current DA methods.

Moreover, the paper delves into the root causes of LLM sparsity, providing a comprehensive theoretical analysis of two key features: history-related activation uncertainty and semantic-irrelevant activation inertia. These insights not only establish a robust theoretical foundation for DA methods but also offer valuable guidance for future research aimed at optimizing LLMs for greater efficiency and effectiveness. Through detailed analyses and empirical validation, this work paves the way for more efficient utilization of LLMs, potentially enhancing their application across various domains.

\section*{Limitations}
This paper highlights that sequence-level activation is predominantly influenced by key elements within the same sequence. However, due to space constraints, ablation studies were not included. The datasets and the volume of data used are limited, necessitating more extensive experiments in future research.

The theoretical derivations indicate that the rich information in sequences can be effectively leveraged to uncover activation patterns and optimize the inference process.

Future work will focus on utilizing sequence information for mixture-of-depth selection, dynamically selecting the appropriate model depth during inference to reduce computational overhead and enhance efficiency. Additionally, compressing the prompt portion can reduce input sequence length and complexity, thereby decreasing generation latency and improving the model's responsiveness and resource utilization efficiency while maintaining performance.

\bibliography{aaai25}

\appendix
\section{Training-Dependent DA in MoEs}\label{app:training-dependent da in moes}
MoE models can achieve high performance with fewer activation parameters. Inspired by this, the DA method adopts a similar structure by converting the FFN layers of dense models into experts and employing multi-stage training to achieve both high performance and sparsity. This approach leverages the model's inherent sparsity, transforming it into sparse activation of the experts.

MoEfication\cite{zhang2022moeficationtransformerfeedforwardlayers} emulates the dynamic and sparse activation of the human brain by transforming FFNs into MoEs. This process is accomplished in two stages: 1) dividing the parameters of the FFNs into multiple experts, and 2) constructing an expert router to determine which experts to use for each input. Experimental results indicate that MoEfication can maintain model performance across various downstream tasks while reducing FFN parameters by 10-30\%.

DS-MoE\cite{pan2024densetrainingsparseinference} introduces a framework that employs dense computation during training and switches to sparse computation during inference. LLaMA-MoE\cite{zhu2024llamamoebuildingmixtureofexpertsllama} offers a new lightweight method to transform FFNs into MoEs. LTE\cite{zheng2024learnefficientbuildstructured} achieves a superior balance between sparsity and performance by activating fewer neurons and is applicable to models with both ReLU and non-ReLU activation functions.
Lory\cite{zhong2024loryfullydifferentiablemixtureofexperts} retains the autoregressive properties of language models by adopting a causally segmented routing strategy and a similarity-based data batching method. This enables efficient expert merging operations and promotes specialization among experts in processing similar documents during training sessions.

\section{Proof of Claim~\ref{claim:cause of sparsity}}\label{app: proof of claim}
Expanding the enumerator in the first term on the RHS of Equation \ref{eq: lianshiqiudao second term} yields Equation \ref{eq: enumerator}:
\begin{figure*}[h]
\begin{equation}
\begin{aligned}
    \left \langle exp(\sum_{i}\sigma(p_i) \cdot \boldsymbol{v_i} ),\boldsymbol{v_{i^*}} \right \rangle 
    &= \sum_{m}(v_{i^*,m} \cdot exp(\sum_{i}\sigma(p_i) \cdot v_{im}) \\
    &= \sum_{m}(v_{i^*,m} \cdot exp(p_{i^*} \cdot v_{i^*m}) \cdot exp(\sum_{i \neq i^*}\sigma(p_i) \cdot v_{im}) 
\end{aligned}
\label{eq: enumerator}
\end{equation} 
\end{figure*}

In Equation\ref{eq: enumerator}, we assume that parameter \( \theta \) and \( \tau \) have no negative features. 

If we have:
\begin{itemize}
    \item \(p_{i^*}^0=Swish_1(x\theta)\odot(x\tau)\), and
    \item \(p_{i^*}^1=ReLU(x)\)
\end{itemize}
respectively, it is easy to get:
\begin{itemize}
    \item \(Swish_1(x\theta) < x\theta\) when \(x>0\), and
    \item \(p_{i^*}^0 < x\theta = p_{i^*}^1\), and
    \item \(p_{i^*}^0 < x\tau\)
\end{itemize}
holds true.

By substituting Equation \ref{eq: enumerator} into Equation \ref{eq: lianshiqiudao second term} and denoting:
\begin{equation}
\begin{aligned}
    C^{(1)}_m=exp(exp(\sum_{i \neq i^\ast}\sigma(p_i) \cdot v_{im})) 
\end{aligned}
\end{equation}
, we then have Equation \ref{eq: c1}:
\begin{equation}\footnotesize
\begin{aligned}
    \frac{\partial \ell_{CE}}{\partial p_{i^*}}
    &=\sum_{m}(\frac{v_{i^\ast,m} \cdot exp(p_{i^\ast} \cdot v_{i^\ast,m}) \cdot C^{(1)}_m}{\left \langle exp(\sum_{i}\sigma (p_i) \cdot \boldsymbol{v_i}), \boldsymbol{1} \right \rangle } )-\left \langle \boldsymbol{v_{i^\ast}}, \boldsymbol{y} \right \rangle 
\end{aligned}
\label{eq: c1}
\end{equation}

For the denominator in the first term on the RHS of the Equation \ref{eq: c1}, we have Equation \ref{eq: c2}:
\begin{figure*}[h]
\begin{equation}
\begin{aligned}
    \left \langle exp(\sum_{i}\sigma (p_i) \cdot \boldsymbol{v_i}), \boldsymbol{1} \right \rangle
    &=\sum_{m^\prime}exp(\sum_{i}\sigma(p_i) \cdot v_{im^\prime}) \\
    &=\sum_{m^\prime}(exp(p_{i^\ast}\cdot v_{i^\ast,m^\prime})\cdot exp(\sum_{i\ne i^\ast}\sigma(p_i) \cdot v_{im^\prime})) \\
    &=exp(p_{i^\ast}\cdot v_{i^\ast,m})\cdot exp(\sum_{i\ne i^\ast}\sigma(p_i) \cdot v_{i,m})+\sum_{m^\prime \ne m}(exp(p_{i^\ast}\cdot v_{i^\ast,m^\prime})\cdot exp(\sum_{i\ne i^\ast}\sigma(p_i) \cdot v_{im^\prime})
\end{aligned}
\label{eq: c2}
\end{equation}
\end{figure*}

By substituting Equation \ref{eq: c2} into Equation \ref{eq: c1} and denoting:
\begin{equation}
\begin{aligned}
    C^{(2)}_m=exp(\sum_{i\ne i^\ast}\sigma(p_i) \cdot v_{im^\prime}))
\end{aligned}
\end{equation}
and
\begin{equation}
\begin{aligned}
    C^{(3)}_m=\sum_{m^\prime \ne m}(exp(p_{i^\ast}\cdot v_{i^\ast,m^\prime})\cdot exp(\sum_{i\ne i^\ast}\sigma(p_i) \cdot v_{im^\prime})
\end{aligned}
\end{equation}
, then we have Equation \ref{eq: c3}:
\begin{equation}\footnotesize
\begin{aligned}
    \frac{\partial \ell_{CE}}{\partial p_{i^*}}
    &=\sum_{m}(\frac{v_{i^\ast,m} \cdot exp(p_{i^\ast} \cdot v_{i^\ast,m}) \cdot C^{(1)}_m}{exp(p_{i^\ast}\cdot v_{i^\ast,m}) \cdot C^{(2)}_m + C^{(3)}_m} )-\left \langle \boldsymbol{v_{i^\ast}}, \boldsymbol{y} \right \rangle
\end{aligned}
\label{eq: c3}
\end{equation}

Taking expectation with respect to all entries of V are independent, we thus can get Equation \ref{eq: loss expectation} in Section \ref{sec: unveiling}.

\section{Larger Scale of Activation Inertia}\label{app: larger scale}

\begin{figure*}[p]
    \begin{minipage}[t]{0.5\linewidth}
        \centering
        \includegraphics[height=2in]{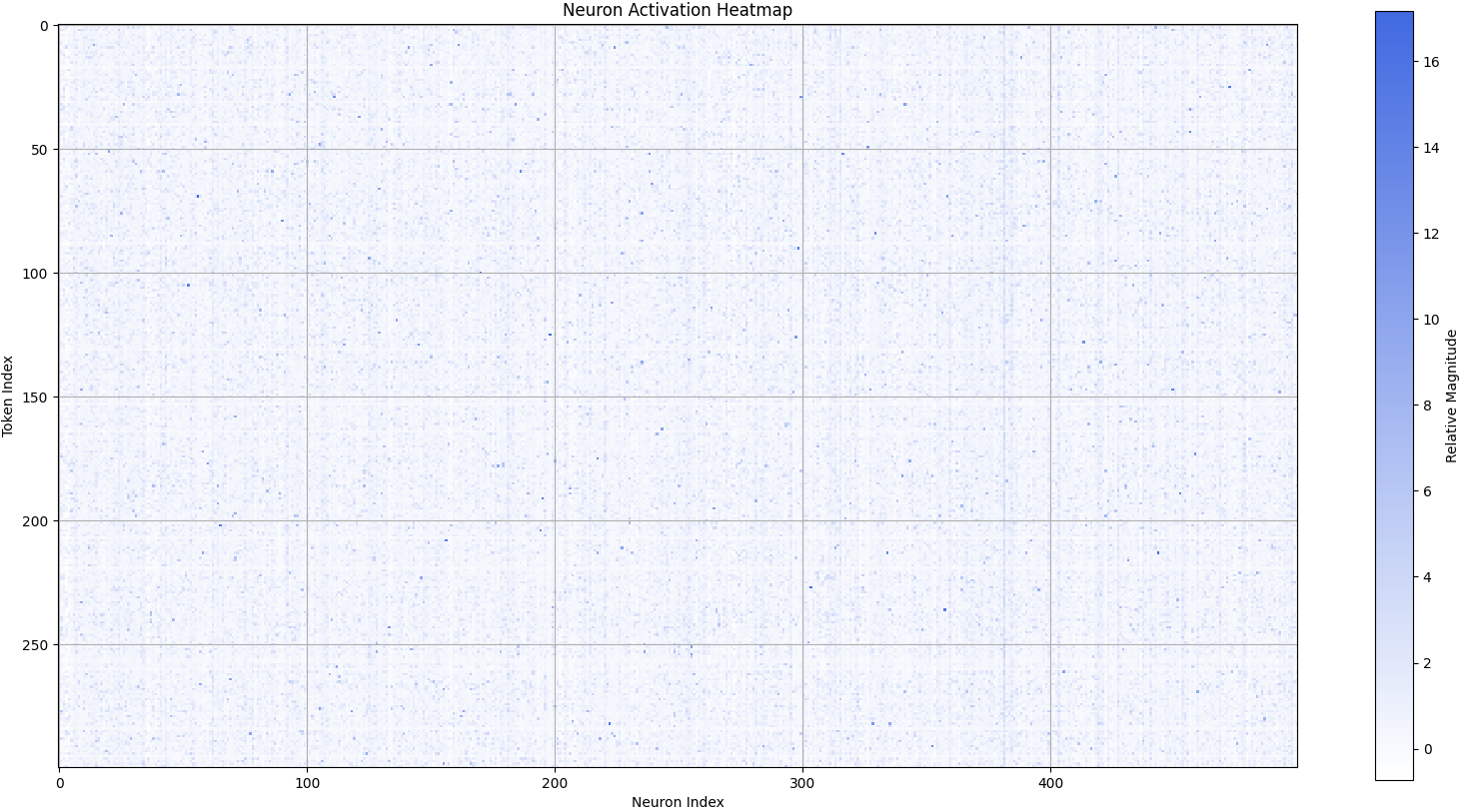}
        \caption{Active neuron of each token from a sentence}
        \label{figure: sent par}
    \end{minipage}
    \begin{minipage}[t]{0.5\linewidth}
        \centering
        \includegraphics[height=2in]{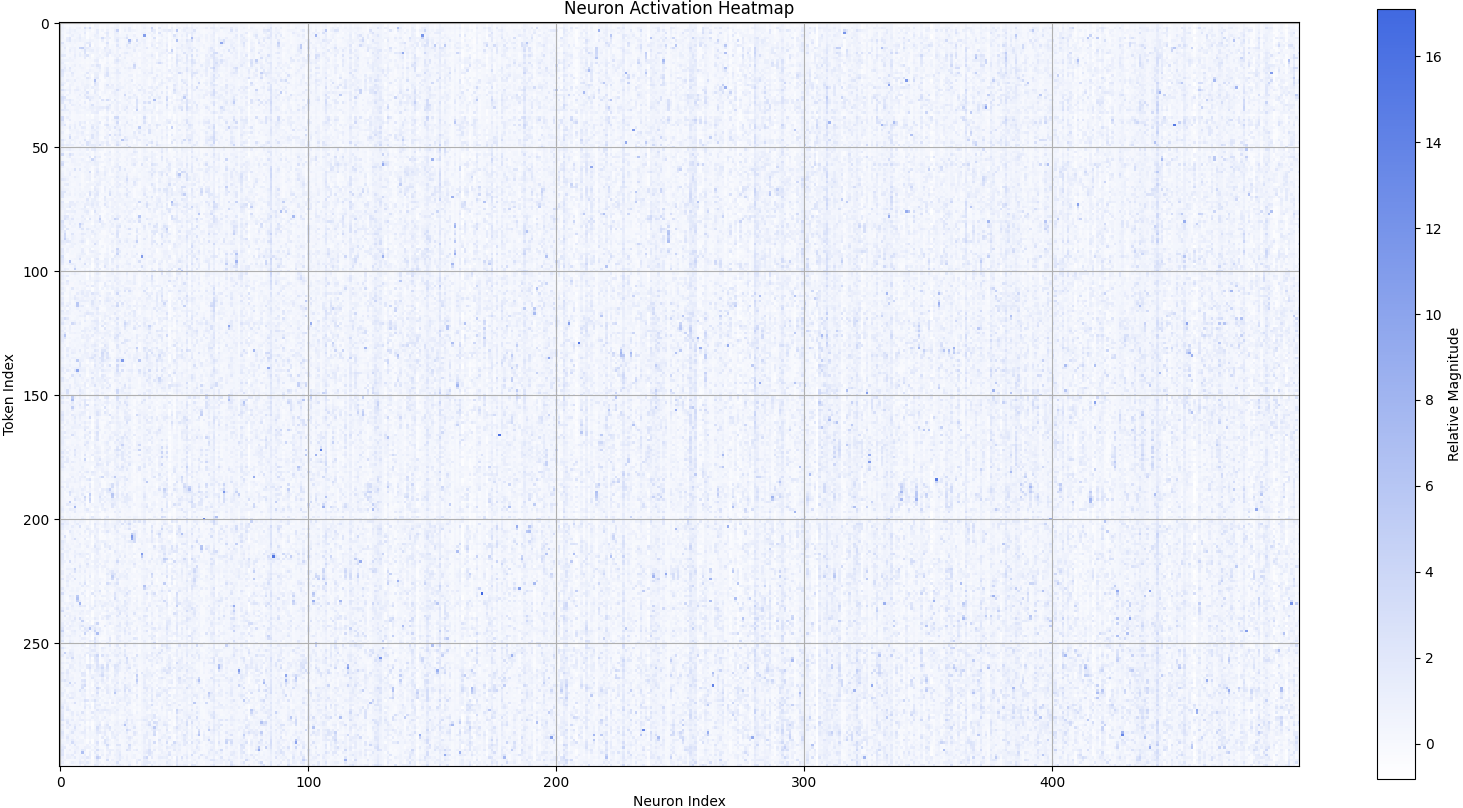}
        \caption{Active neuron of this sentence}
        \label{figure: sent seq}
    \end{minipage}
        \begin{minipage}[t]{0.5\linewidth}
        \centering
        \includegraphics[height=2in]{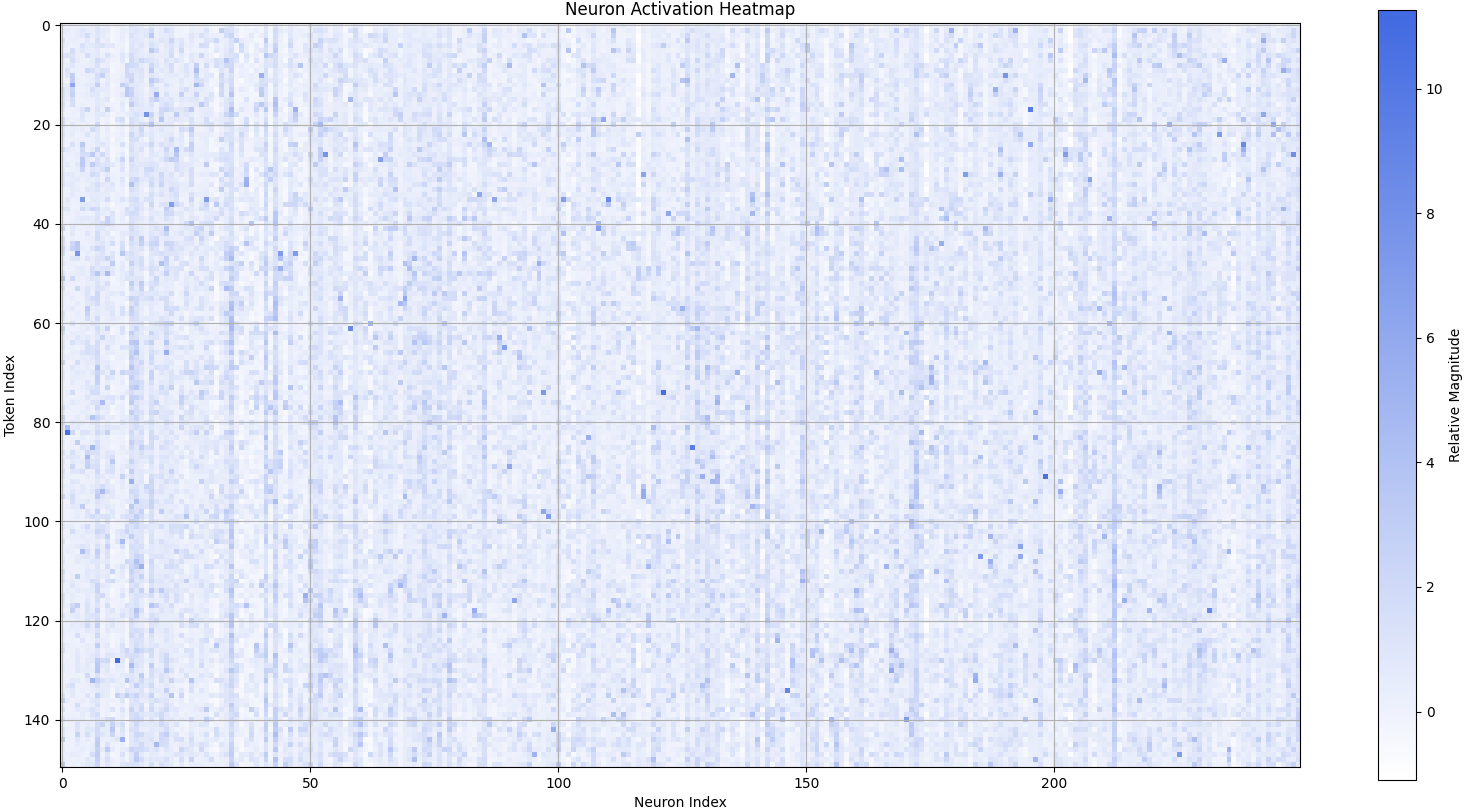}
        \caption{Active neuron of each random token}
        \label{figure: rand par}
    \end{minipage}
        \begin{minipage}[t]{0.5\linewidth}
        \centering
        \includegraphics[height=2in]{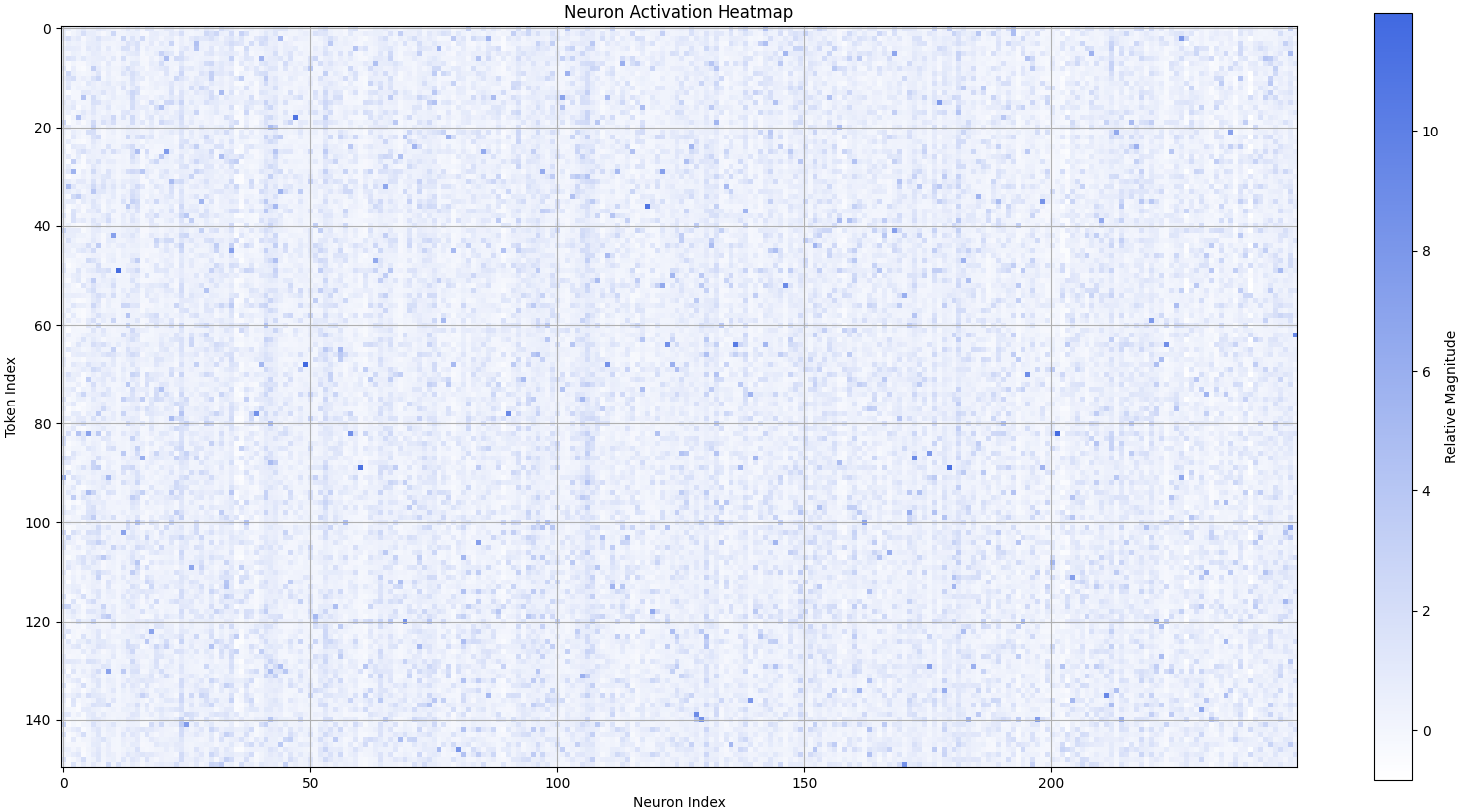}
        \caption{Active neuron of these random tokens\protect\\ as a sequence}
        \label{figure: rand seq}
    \end{minipage}
\end{figure*}

Figure \ref{figure: sent par} to Figure \ref{figure: rand seq} demonstrate the existence of activation inertia and its irrelevance to semantics. The horizontal axis represents different neurons, while the vertical axis represents different samples. Zoom in for better experience.

From Figure \ref{figure: sent par} to Figure \ref{figure: rand seq}, we can draw the following conclusions:
\begin{enumerate}
    \item Figures \ref{figure: sent par} and Figures \ref{figure: sent seq} illustrate the active neurons for tokens from a \textit{sentence}, either individually or as part of the sentence. Although neither is very pronounced, Figure \ref{figure: sent seq} has more noticeable blue stripes compared to Figure \ref{figure: sent par}.
    \item Conversely, Figures \ref{figure: rand par} and \ref{figure: rand seq} display the active neurons when tokens from a \textit{random word list} are processed either individually or as part of the sentence. Similarly, Figure \ref{figure: rand seq} has more noticeable blue stripes compared to Figure \ref{figure: rand par}.
\end{enumerate}

Therefore, during sequential input, neuronal activation becomes more flocking. Additionally, random words tend to intensify this trend of concentrated activation. These two conclusions are consistent with Griffin\cite{dong2024promptpromptedadaptivestructuredpruning}.

\section{Samples for activation inertia check}\label{app: detailed samples}

\begin{table*}[p]
\centering
\scalebox{0.9}{
\begin{tabular}{c|ll}
\toprule
\textbf{Index} & \multicolumn{1}{c}{\textbf{Samples}}                               & \multicolumn{1}{c}{\textbf{Treatments}}          \\
\midrule
1              & "\#\#\# Article: Almost one million people visited the city"       & Baseline                                   \\
2              & "Article: Almost one million people visited the city"              & Remove beginning token                     \\
3              & "Almost one million people visited the city"                       & Remove beginning tokens                    \\
4  & "\#\#\# Article: Nearly one million people visited the city" & Modify the word at the beginning of the sequence. \\
5              & "Nearly one million people visited the city"                       & Remove beginning tokens                    \\
6              & "\#\#\# Article: Less than one million people visited the city"    & Change to antonym                          \\
7              & "Less than one million people visited the city"                    & Remove beginning tokens                    \\
8              & "\#\#\# Article: Almost one million people visited the city"       & Similarity threshold                          \\
9              & "\#\#\# Article: Almost one million people visited the restaurant" & Change to synonyms                        \\
10 & "Almost one million people visited the restaurant"           & Modify the word at the end of the sequence        \\
11             & "Almost one million people visited the planet"                     & Modify the word at the end of the sequence \\
12 & "Almost one million tourists visited the restaurant"         & Modify the words at the middle and end            \\
13             & "Almost one million aliens visited the planet"                     & Dissimilarity threshold    \\        
\bottomrule
\end{tabular}
}
\caption{Detailed 13 samples for activation inertia check.}
\label{table: detailed 13 samples}
\end{table*}

Table~\ref{table: detailed 13 samples} details the 13 samples used for activation pattern similarity analysis. Samples 1-3 and Samples 4, 6, and 9 form two treatment groups. If Sample 4 shows greater similarity to Sample 1 than to Samples 2 and 3, it supports Claim~\ref{final claim}.

\begin{figure*}[p]
\centering
\includegraphics[scale=0.8]{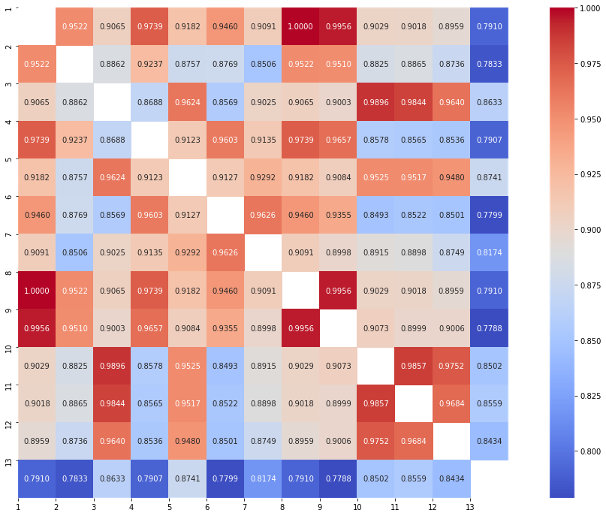}
\caption{Similarity matrix of 13 samples' activation pattern}
\label{figure: similarity}
\end{figure*}

From the similarity heatmap in Figure~\ref{figure: similarity}, we observe the following:
a) Samples 4, 6, and 9 are more similarly activated to Sample 1 than to Samples 2 and 3;
b) Samples 1, 6, 8, and 9 are more similarly activated to Sample 4 than to Sample 5;
c) Sample 9 is more similarly activated to Samples 4, 6, and 8;
d) Samples 11 and 12 are more similarly activated to Samples 9 and 10;
e) Samples 10, 11, and 12 are more similarly activated to Sample 13 than to any other samples.

\section{Layer-wise Threshold and CETT}\label{app: cett}
The formula for LLaMA's MLP block can be described in Equation \ref{eq: MLP block} given an input x:
\begin{equation}
    MLP(x)=W^{out}\left [ \sigma(W^{in}x)\odot (V^{in}x) \right ] 
\label{eq: MLP block}
\end{equation}
, where the output of the i-th neuron can be defined as Equation \ref{eq: ith neuron}:
\begin{equation}
    n_i(x)=\left [ \sigma (W^{in}_{i,:}x)\odot (V^{in}_{i,:}x) \right ]W^{out}_{:,i} 
\label{eq: ith neuron}
\end{equation}

From Equation \ref{eq: MLP block} and Equation \ref{eq: ith neuron}, it can be easily obtained that (Equation \ref{eq: MLP sum from neuron}):
\begin{equation}
    MLP(x)= {\textstyle \sum_{i=1}^{d_{h}}n_i(x)} 
\label{eq: MLP sum from neuron}
\end{equation}
, where $d_{h}$ is the dimension of the hidden layer in MLP block. Therefore, the formula for CETT(cumulative errors of tail truncation) is as follows in Equation \ref{eq: mlp cett}:
\begin{equation}
\begin{aligned}
    CETT(x)&=\frac{|| {\textstyle \sum_{i\in \mathcal D}n_i(x)} ||_2}{||MLP(x)||_2},\\
    \mathcal D&=\left \{i|\;||n_i(x)||_2<\epsilon \right \} 
\end{aligned}
\label{eq: mlp cett}
\end{equation} 
, where \(\epsilon \) represents the threshold, $\mathcal D$ is the set of neurons with magnitudes less than the threshold \(\epsilon \), and $n_i$ denotes the output of the i-th neuron from Equation \ref{eq: ith neuron}. Generally, the CETT is empirically set at 0.2, after which the maximum \(\epsilon \) achievable is calculated to determine the threshold.
\end{document}